\documentclass[11pt]{article}

\usepackage[preprint]{acl}

\usepackage{times}
\usepackage{latexsym}
\usepackage{amsfonts}
\usepackage{amsmath}
\usepackage{subcaption}
\usepackage{graphicx}
\usepackage{scalefnt} 

\usepackage{graphicx}
\usepackage{xcolor}
\usepackage{CJKutf8}
\usepackage{tabularx}
\usepackage{booktabs}
\usepackage{makecell}
\usepackage[misc]{ifsym}
\usepackage{amssymb}
\usepackage{pifont}
\usepackage[export]{adjustbox}
\usepackage{multirow}
\usepackage{diagbox}
\usepackage{bm}

\newcommand{\gray}[1]{\textcolor{gray}{#1}}

\usepackage[T1]{fontenc}

\usepackage[utf8]{inputenc}

\usepackage{microtype}

\usepackage{inconsolata}

\usepackage{graphicx}

\title{Efficient Low-Resource Language Adaptation\\via Multi-Source Dynamic Logit Fusion}

\author{Chen Zhang, Jiuheng Lin, Zhiyuan Liao, Yansong Feng\thanks{Corresponding author.} \\
Wangxuan Institute of Computer Technology, Peking University \\
{\tt \{zhangch,fengyansong\}@pku.edu.cn}\\
{\tt \{linjiuheng,liaozy\}@stu.pku.edu.cn}
}

\begin{document}
\maketitle
\begin{abstract}
Adapting large language models (LLMs) to low-resource languages (LRLs) is constrained by the scarcity of task data and computational resources. Although Proxy Tuning offers a logit-level strategy for introducing scaling effects, it often fails in LRL settings because the large model’s weak LRL competence might overwhelm the knowledge of specialized smaller models. We thus propose \textsc{TriMix}, a test-time logit fusion framework that dynamically balances capabilities from three different sources: LRL competence from a continually pretrained small model, task competence from high-resource language instruction tuning, and the scaling benefits of large models. It is data- and compute-efficient, requiring no LRL task annotations, and only continual pretraining on a small model. Experiments across four model families and eight LRLs show that \textsc{TriMix} consistently outperforms single-model baselines and Proxy Tuning. Our analysis reveals that prioritizing the small LRL-specialized model's logits is crucial for success, challenging the prevalent large-model-dominant assumption.
\end{abstract}

\section{Introduction}
Although large language models (LLMs) have achieved remarkable success in high-resource languages (HRLs), their ability to process low-resource languages (LRLs) remains limited~\cite{singh-etal-2025-global}.
To improve the LLM performance in LRLs, researchers attempt to adapt the HRL-dominant LLMs to LRLs~\cite{ke-etal-2025-adaptation}, which typically face two fundamental challenges.  
One challenge is the scarcity of labeled task data for these languages~\cite{joshi-etal-2020-state}, due to the high cost of manual annotation. Consequently, there is insufficient fine-tuning data to endow models with task-solving capabilities.
The other challenge lies in the scarcity of computational budget~\cite {urbizu-etal-2025-sub,hernandez2025lessons}. Researchers working on LRLs often rely on smaller models and lack the resources required to continually pretrain large-scale models, even though larger models typically exhibit stronger general abilities.

\begin{figure}[t]
\centering
\includegraphics[width=\linewidth]{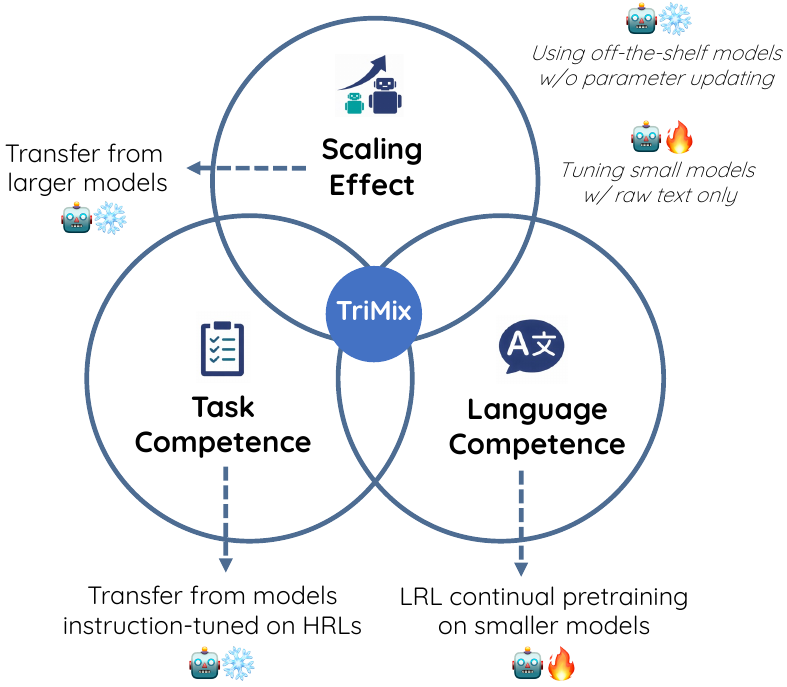}
\caption{\textsc{TriMix} integrates three sources of benefit for LRL adaptation while minimizing the need for annotating task data and tuning larger models.  }
\label{fig:teaser}
\end{figure}

Recent studies have explored paradigms avoiding reliance on annotated LRL task data, such as model merging~\cite{tao-etal-2024-unlocking,yamaguchi2025adapting}. It combines task-solving capabilities learned from HRLs with language competence acquired through continual pretraining (CPT) on the target LRL.
However, model merging requires the merging models to share the same architecture and scale; consequently, scaling up still necessitates CPT on a larger model, which is computationally expensive. 
To alleviate computational constraints, Proxy Tuning~\cite{liu2024tuning} aims to approximate the benefits of training a larger specialized model by injecting the knowledge of a smaller, domain-adapted model into the logits of a larger one. This approach shows promising results in the code domain.
However, applying it to LRL settings is non-trivial. In such scenarios, the larger model itself often lacks sufficient language competence in the target LRL.
As a result, when the larger model dominates the logit arithmetic, as implicitly assumed in Proxy Tuning~\cite{liu2024tuning,zhao2024weaktostrong,zhang2025logit}, its weak LRL representations can overwhelm the contribution of the smaller CPT model, limiting effective transfer and even destroying basic LRL generation ability (see an example in Appendix~\ref{app:case_sudy}).
This observation highlights an important issue in logit-level fusion: different sources of model capability are not always equal and need more careful balancing.

To address these limitations, we propose \textsc{TriMix}, a test-time logit fusion framework integrating (i) language-specific competence acquired through CPT on LRLs, (ii) task competence learned from HRLs, and (iii) the scaling benefits of large models, as shown in Figure~\ref{fig:teaser}.
\textsc{TriMix} requires CPT only on a small model using raw LRL text, without the need of task-level annotation. At inference time, it leverages an off-the-shelf large instruction-tuned model to benefit from scaling effects.
Unlike prior Proxy Tuning approaches that treat the large model as the dominant component by default, \textsc{TriMix} explores adaptive weighting strategies to dynamically balance these different sources of capability.

We validate our framework on four LLM families across eight LRLs. \textsc{TriMix}, particularly when combined with the perplexity-guided weighting strategy, consistently outperforms the single-model baselines as well as Proxy Tuning.
For instance, continually pretraining Qwen2.5-1.5B on LRL corpora and fusing it with a 14B instruction-tuned model yields an average relative improvement of approximately 5\% over the 14B model.

To better understand the mechanisms underlying the effectiveness of \textsc{TriMix}, we analyze the weights assigned during logit fusion. Under the empirical upper bound of \textsc{TriMix}, we find that substantially larger weights are assigned to the small CPT model than to the large instruction-tuned model, in sharp contrast to Proxy Tuning, which typically prioritizes the larger model. Our perplexity-guided strategy for hyperparameter selection closely approximates the behavior of the upper bound setting, explaining its strong empirical performance. Furthermore, we show that divergence from the base model offers a plausible explanation for when language-specific competence should be emphasized.

Our contributions are summarized as follows:
(1) We propose \textsc{TriMix}, an efficient test-time logit fusion framework for LRL, which dynamically integrates language-specific competence and task competence while leveraging the scaling benefits of large models. (2) We validate \textsc{TriMix} across three LLM families, covering multiple small–large model scale pairs and eight LRLs, demonstrating its flexibility and strong generalizability. (3) We challenge the large-model-dominant assumption underlying Proxy Tuning, showing the importance of prioritizing LRL competence in logit fusion.\footnote{Our code are publicly available at \url{https://github.com/luciusssss/TriMix}.}

\begin{figure*}[t]
\centering
\includegraphics[width=\linewidth]{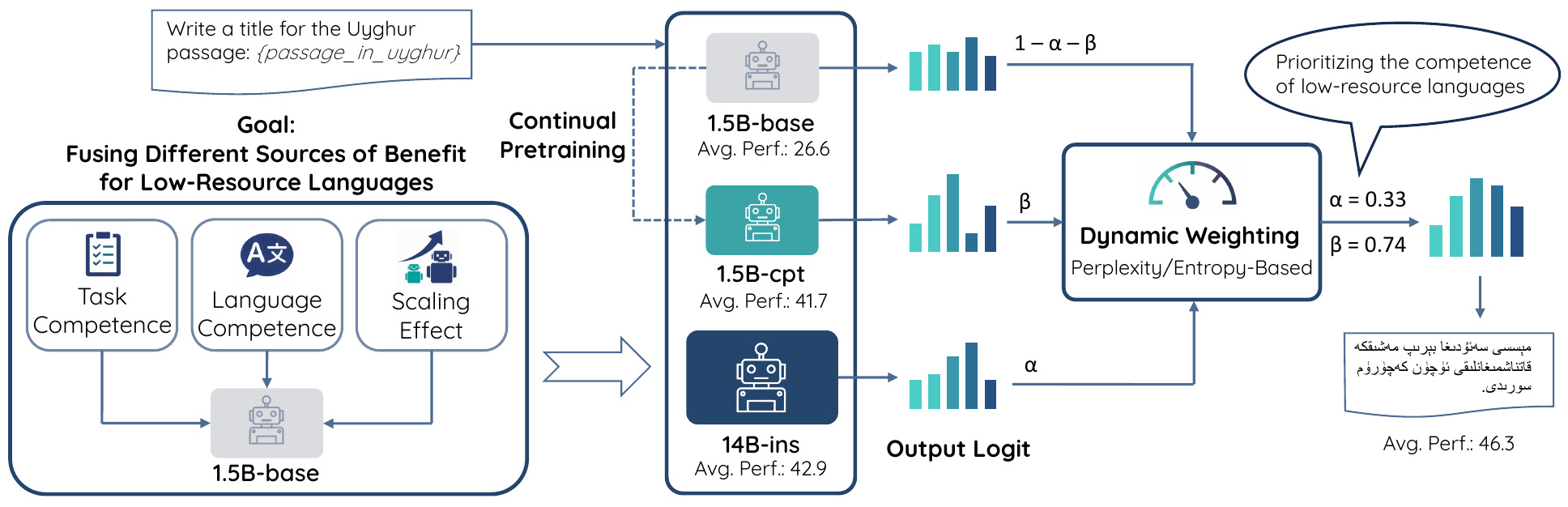}
\caption{The framework of \textsc{TriMix}. Given a task prompt for an LRL, \textsc{TriMix} dynamically fuses the logits of three models to integrate language competence, task competence, and scaling benefits.}
\label{fig:method}
\end{figure*}

\section{Method}
We propose \textsc{TriMix}, a logit fusion framework designed to address the scarcity of annotated task data in LRLs while minimizing the cost of scaling model sizes. 
As illustrated in Figure~\ref{fig:method}, based on the logit arithmetic of three models, \textsc{TriMix} explicitly disentangles and recombines three sources of benefit: (i) language-specific competence acquired via CPT on LRL, (ii) task-solving capabilities transferred from HRLs, and (iii) general capability gains driven by scaling effects.
To effectively integrate these different signals, we introduce an adaptive fusion mechanism that balances their contributions at inference time.

\subsection{Preliminary}
\paragraph{Logit} 
Let $L \in \mathbb{R}^{|V|}$ denote the logit vector (the unnormalized token distribution) produced by a language model before the softmax operation, where $|V|$ is the vocabulary size of the model. Different models (or variants thereof) induce distinct logit distributions reflecting their abilities obtained through specific training data. We follow prior research, such as Proxy Tuning~\cite{liu2024tuning}, that demonstrates that linearly combining logits from different models can effectively synthesize capabilities without further training.

\paragraph{Model Variants}
Our framework leverages open-source LLMs trained primarily on HRLs. We define three key variants:
\begin{itemize}
\item \textbf{Base (base)}: A foundation model pretrained on massive HRL corpora, exhibiting general reasoning and language modeling capabilities.
\item \textbf{Instructed (ins)}: An instruction-tuned variant obtained by fine-tuning the base model on HRL instruction-following data.
\item \textbf{Continually Pretrained (cpt)}: A variant obtained by performing CPT on the base model using LRL corpora to enhance target language competence.
\end{itemize}

To minimize computational overhead, we perform CPT solely on a small model variant (small-cpt) and transfer this competence to a larger model variant (large-ins) via logit fusion.

\subsection{Tri-Source Fusion}

We formulate the fusion objective as a linear decomposition of capabilities. 
We posit that the ideal logit $L$ can be constructed by augmenting a baseline small model with specific benefit vectors:
\begin{equation}
  \label{eq:decomposition}
  L = L_{\text{small-base}} + \alpha \delta_{\text{T}} + \beta \delta_{\text{L}} + \gamma \delta_{\text{S}},
\end{equation}
\noindent where $\delta_{\text{T}}$, $\delta_{\text{L}}$, and $\delta_{\text{S}}$ represent the benefit vectors for task solving, LRL modeling, and scaling, respectively.

\paragraph{Task Solving Vector ($\delta_{\text{T}}$)} 
We isolate the task-solving capability by contrasting the large instruction-tuned model with its base counterpart. We utilize the large model here instead of the small one, as larger models typically exhibit stronger learning capacity under the same amount of data~\cite{kaplan2020scaling}:
\begin{equation}
  \label{eq:instruction}
  \delta_{\text{T}} = L_{\text{large-ins}} - L_{\text{large-base}}.
\end{equation}

\paragraph{Language Modeling Vector ($\delta_{\text{L}}$)}
Due to computational constraints, we assume that CPT is feasible only for the smaller model. Thus, we define the LRL competence vector as:
\begin{equation}
  \label{eq:language}
  \delta_{\text{L}} = L_{\text{small-cpt}} - L_{\text{small-base}}.
\end{equation}

\paragraph{Scaling Effect Vector ($\delta_{\text{S}}$)}
To capture the benefits of scaling, we compute the difference between the large and small base models. One could subtract either the base versions or the instructed versions between the large and small models. We choose the base models to avoid potential confounding effects introduced by instruction tuning:
\begin{equation}
  \label{eq:scale}
  \delta_{\text{S}} = L_{\text{large-base}} - L_{\text{small-base}}.
\end{equation}

To streamline the inference process and reduce memory overhead, we strategically set the scaling coefficient $\gamma = \alpha$. This constraint allows the $L_{\text{large-base}}$ terms in Eq.~\ref{eq:instruction} and Eq.~\ref{eq:scale} to cancel out, eliminating the need to load the large base model into memory. As a result, both the task-solving capability and the scaling effect are injected through a single large instruction-tuned model\footnote{This constraint trades modeling flexibility for efficiency: while allowing $\gamma \neq \alpha$ might yield better performance, we adopt $\gamma = \alpha$ as a practical approximation and leave unconstrained fusion to future work.}.
Substituting the definitions into Eq.~\ref{eq:decomposition} yields our final formulation:
\begin{equation}
\label{eq:final}
\begin{split}
L = &\ \alpha L_{\text{large-ins}} + \beta L_{\text{small-cpt}} \\
    &\ + (1 - \alpha - \beta)\, L_{\text{small-base}}.
\end{split}
\end{equation}

\subsection{Dynamic Weighting}
\label{sec:hyperparameter}
In the final formulation (Eq.~\ref{eq:final}), two hyperparameters must be determined. 
In LRL settings, we typically lack sufficient annotated data to perform an extensive hyperparameter search. 
Consequently, we propose two heuristics to determine these weights dynamically at inference time. 

\paragraph{Perplexity-Guided Selection (PPL)} 
The perplexity of the input prompt serves as a proxy for how well the fused model captures the input distribution~\cite{mavromatis2024pack,xu-etal-2025-hit}. We select the ($\alpha$, $\beta$) pair that minimizes the perplexity of the  prompt, which consists of in-context learning examples and the input of the current test instance.

\paragraph{Entropy-Guided Selection (ENT)} 
Alternatively, we utilize predictive confidence as a selection metric. Inspired by previous works~\cite{garces-arias-etal-2024-adaptive,jin-etal-2024-dvd}, we calculate the entropy of the next-token distribution. We select the hyperparameters that minimize the entropy of the first generated token, prioritizing configurations where the model exhibits high certainty.

\begin{table}[t]
\setlength\tabcolsep{4pt}
\centering
\begin{small}
\begin{tabular}{lccc}
\toprule
\textbf{Family} & \textbf{small-cpt} & \textbf{large-ins} & \textbf{Languages} \\
\midrule
Qwen2.5 & 1.5B/3B/7B & 3B/7B/14B & bod, uig, kaz, mvf \\
\midrule
\multirow{2}{*}{Llama2}  & \multirow{2}{*}{7B} & \multirow{2}{*}{13B} & bod, uig, mvf,  \\
& & & tam, tel, ory, ben \\
\midrule
Llama3.2 & 1B & 3B & bod, uig, kaz \\
\midrule
Gemma3 & 4B & 12B & bod, uig, kaz, mvf \\
\bottomrule
\end{tabular}
\end{small}
\caption{The model scales and evaluated languages of each model family.}
\label{tab:family}
\end{table}

\begin{table*}[t]
\setlength\tabcolsep{5pt}
\centering
\begin{small}
\begin{tabular}{l|cc|ccc|ccccl}
\toprule
\multirow{2}{*}{\textbf{Method}} & \textbf{\#Param} & \textbf{\#Param}  & \multirow{2}{*}{\textbf{MC}} & \multirow{2}{*}{\textbf{ENG-G}} & \multirow{2}{*}{\textbf{LRL-G}} & \multirow{2}{*}{\textbf{bod}} & \multirow{2}{*}{\textbf{uig}} & \multirow{2}{*}{\textbf{kaz}} & \multirow{2}{*}{\textbf{mvf}} & \multirow{2}{*}{\textbf{Average}} \\
& \textbf{Train} & \textbf{Test} & & & & & & \\
\midrule
\multicolumn{11}{c}{small = 1.5B, large = 1.5B} \\
\midrule
Qwen2.5-1.5B-base$^{\dag}$ & 0B & 1.5B & 40.2 & 10.8 & 11.5 & 22.3 & 26.6 & 25.5 & 20.0 & 23.6 \\
Qwen2.5-1.5B-ins$^{\dag}$ & 0B & 1.5B & 36.4 & 10.4 & 10.5 & 19.8 & 24.3 & 23.4 & 18.9 & 21.6  \\
Qwen2.5-1.5B-cpt & 1.5B & 1.5B & \textbf{48.3} & \underline{17.6} & \underline{17.3} & \underline{25.7} & \underline{41.7} & \textbf{33.8} & \textbf{21.5} & 30.7  \\
Contrastive Decoding & 1.5B & 3B & 45.3 & 15.7 & 15.2 & 25.0 & 37.7 & 32.1 & 18.1 & 28.2 \tiny{(-8.1\%)} \\
Model Merging & 1.5B & 1.5B & \underline{48.0} & \textbf{18.6} & \textbf{17.5} & \textbf{27.0} & \textbf{41.8} & \textbf{33.8} & \underline{21.0} & \textbf{30.9} \tiny{(+0.7\%)} \\
\midrule
\multicolumn{11}{c}{small = 1.5B, large = 3B} \\
\midrule
Qwen2.5-3B-ins$^{\dag}$ & 0B & 3B & 42.4 & 12.2 & 10.8 & \underline{24.2} & 30.4 & 23.2 & \underline{21.2} & 24.8  \\
Proxy Tuning & 1.5B & 6B & \underline{45.4} & \underline{14.1} & 14.1  & 23.5 & \underline{40.0} & 29.4 & \underline{21.2} & \underline{28.5} \tiny{(-7.2\%)}\\
\textsc{TriMix} (ENT) & 1.5B & 6B & 45.0 & 13.7 & \underline{14.8}  & 24.1 & 33.8 & 29.3 & \textbf{22.5} & 27.4 \tiny{(-10.7\%)}\\
\textsc{TriMix} (PPL) & 1.5B & 6B & \textbf{48.7} & \textbf{19.5} & \textbf{16.3} & \textbf{25.6} & \textbf{41.4} & \textbf{36.4} & 21.1 & \textbf{31.1} \tiny{(+1.3\%)} \\
\gray{\textit{\textsc{TriMix} (Upper Bound)}} & \gray{\textit{1.5B}} & \gray{\textit{6B}} & \gray{\textit{52.4}} & \gray{\textit{21.3}} & \gray{\textit{17.6}} & \gray{\textit{27.9}} & \gray{\textit{43.8}} & \gray{\textit{37.4}} & \gray{\textit{25.3}} & \gray{\textit{33.6} \tiny{(+9.4\%)}}  \\
\midrule
\multicolumn{11}{c}{small = 1.5B, large = 7B} \\
\midrule
Qwen2.5-7B-ins$^{\dag}$ & 0B & 7B & 49.7 & \textbf{20.0} & 12.5 & \textbf{28.2} & \underline{40.6} & 30.5 & \underline{23.0} & 30.6  \\
Proxy Tuning & 1.5B & 10B & 50.5 & 16.3 & 13.3 & 26.3 & 38.6 & 33.7 & 21.6 & 30.0 \tiny{(-2.3\%)} \\
\textsc{TriMix} (ENT) & 1.5B & 10B & \underline{51.2} & 16.1 & \textbf{16.4} & 26.7 & 40.5 & \underline{34.8} & 22.9 & \underline{31.2} \tiny{(+1.6\%)} \\
\textsc{TriMix} (PPL) & 1.5B & 10B & \textbf{53.4} & \underline{19.8} & \underline{15.7} & \underline{27.7} & \textbf{42.7} & \textbf{38.1} & \textbf{23.6} & \textbf{33.0} \tiny{(+7.5\%)} \\
\midrule
\multicolumn{11}{c}{small = 1.5B, large = 14B} \\
\midrule
Qwen2.5-14B-ins$^{\dag}$ & 0B & 14B & 57.1 & \textbf{21.0} & 13.8 & \textbf{35.7} & \underline{42.9} & 34.6 & 24.6 & \underline{34.4}  \\
Proxy Tuning & 1.5B & 17B & 57.7 & 15.4 & \underline{16.8} & 33.1 & 40.3 & \underline{37.3} & 25.0 & 33.9 \tiny{(-1.5\%)} \\
\textsc{TriMix} (ENT) & 1.5B & 17B & \underline{57.8} & 15.6 & \textbf{17.0} & 33.3 & 41.3 & 35.8 & \underline{25.8} & 34.1 \tiny{(-0.9\%)} \\
\textsc{TriMix} (PPL) & 1.5B & 17B & \textbf{59.5} & \underline{20.5} & \underline{16.8} & \underline{33.6} & \textbf{46.3} & \textbf{38.6} & \textbf{26.0} & \textbf{36.1} \tiny{(+4.9\%)} \\
\bottomrule
\end{tabular}
\end{small}
\caption{Evaluation results on Qwen2.5 models with different model size combinations. $^{\dag}$ denotes off-the-shelf single models. \textbf{\#Param Train} and \textbf{\#Param Test} denote the number of parameters updated during CPT and used during inference, respectively. \textbf{Bold} indicates the best result, and \underline{underlined} indicates the second best. The numbers in parentheses in the \textbf{Average} column represent relative improvements over the best single-model baselines.}
\label{tab:main_experiment}
\end{table*}

\section{Experiments}
\subsection{Experimental Setups}

\paragraph{Models}
We primarily evaluate our framework using the different scales of models from \textbf{Qwen2.5} family~\cite{yang2024qwen2}.  
To assess the generalizability across architectures, we also conduct experiments with the \textbf{Llama2}~\cite{touvron2023llama}, \textbf{Llama3.2}~\cite{grattafiori2024llama}, and \textbf{Gemma3}~\cite{team2025gemma} series, which exhibit different levels of multilingual abilities. 
We choose these model families because they offer different range of model sizes, which is required for our experiments.
For Llama2, we directly use the CPT checkpoints from \citet{tao-etal-2024-unlocking}; for other model series, we continually pretrain the base models as described in Appendix~\ref{app:cpt}.

\paragraph{Languages}
We focus on eight LRLs spanning diverse linguistic families and scripts.
See their linguistic details in Appendix~\ref{app:languages}.
We mainly evaluate on four minority languages in China, including Tibetan (bod), Uyghur (uig), Kazakh (kaz, Arabic script), and Mongolian (mvf, traditional script).
For Llama2, we additionally evaluate on four Indian languages using the CPT checkpoints from \citet{tao-etal-2024-unlocking}, including Tamil (tam), Telugu (tel), Odia (ory), and Bengali (ben).
In Table~\ref{tab:family}, we summarize the model scales and evaluated languages\footnote{For Llama2, the publicly available CPT checkpoints from \citet{tao-etal-2024-unlocking} do not cover Kazakh. For Llama3.2, we observe no noticeable improvement after CPT on Mongolian, likely due to inadequate tokenizer support and the limited capacity of the smaller model; therefore, we exclude this language from our experiments.}.

\paragraph{Evaluation Datasets}
For the four minority languages in China, we adopt the \textbf{MiLiC-Eval} benchmark~\cite{zhang-etal-2025-milic}. We group its seven tasks into three categories: \textbf{(1) Multi-Choice (MC):} topic classification, response selection, and reading comprehension; \textbf{ (2) Generation in English (ENG-G):} LRL-to-English translation and mathematical reasoning (with English Chain-of-Thought); \textbf{(3) Generation in LRL (LRL-G):} title generation and English-to-LRL translation.
For the Indian languages, following \citet{tao-etal-2024-unlocking}, we evaluate on the \textbf{Belebele} reading comprehension dataset~\cite{bandarkar-etal-2024-belebele} and the \textbf{SIB-200} topic classification dataset~\cite{adelani-etal-2024-sib}.

\paragraph{Baseline Methods}
We compare \textsc{TriMix} against established methods for capability transfer and logit manipulation.

\textbf{Model Merging}~\cite{tao-etal-2024-unlocking,huang-etal-2024-chat,akiba2025evolutionary} combines homogeneous models with complementary capabilities by merging their parameters. In our setting, we merge a small-cpt model and a small-ins model to transfer task-solving capabilities learned from HRLs to LRLs. Following \citet{tao-etal-2024-unlocking}, we adopt the widely used TIES algorithm~\cite{yadav2023tiesmerging} for merging.

\textbf{Contrastive Decoding}~\cite{li-etal-2023-contrastive} amplifies the signal of a \textit{strong} model by subtracting the logits of a \textit{weak} model. In the LRL setting, we treat small-cpt as the strong model and small-base as the weak model to enhance language competence:
\begin{equation}
\label{eq:contrastive}
L = L_{\text{small-cpt}} + \beta (L_{\text{small-cpt}} - L_{\text{small-base}}).
\end{equation}

\textbf{Proxy Tuning}~\cite{liu2024tuning} attempts to transfer the knowledge learned in a smaller model to a larger one by logit arithmetic.
It constitutes a special case of our \textsc{TriMix} formulation where the instruction weight is fixed at $\alpha=1$.
The decoding objective becomes:
\begin{equation}
\label{eq:pt}
L = L_{\text{large-ins}} +  \beta(L_{\text{small-cpt}} - L_{\text{small-base}}).
\end{equation}
\citet{liu2024tuning} originally set $\beta$ to 1 for simplicity, which performs well in practice.
Consistent with this choice, our experiments, such as the hyperparameter analysis in Figure~\ref{fig:heatmap}, indicate that $\beta=1$ generally yields better performance than smaller values when $\alpha=1$.

\textbf{Upper Bound:} To estimate the performance ceiling of \textsc{TriMix}, we conduct an exhaustive grid search over $\alpha$ and $\beta$ in Eq.~\ref{eq:final} and report the optimal achievable performance for each task. Due to the substantial computational cost of evaluating 49 hyperparameter pairs across 28 tasks, we restrict this oracle search to the setting of Qwen2.5 1.5B-small + 3B-ins.

\paragraph{Implementation Details}
For hyperparameter selection in Sec.~\ref{sec:hyperparameter}, we perform an efficient grid search over the discrete set $\{0.1, 0.3, 0.5, 0.7, 0.9, 1.0\}$ for both $\alpha$ and $\beta$~\footnote{As discussed in Appendix~\ref{app:hyperparam_search}, we find that model performance degrades when the scales of logits from different models differ substantially; therefore, we restrict the search to this range to avoid excessive imbalance during fusion.}. 
We estimate a set of optimal fusion hyperparameters for each task using a subsample of 50 examples.
See Appendix~\ref{app:implementation} for implementation details of pretraining and inference.

\begin{table*}[t]
\centering
\begin{small}
\begin{tabular}{l|ccccccc|l}
\toprule
\textbf{Model} & \textbf{bod} & \textbf{uig} & \textbf{mvf} & \textbf{tam} & \textbf{tel} & \textbf{ory} & \textbf{ben} & \textbf{Average} \\
\midrule
7B-base & 17.0 & 18.3 & 12.9 & 26.4 & 21.8 & 19.3 & 31.7 & 21.1\\
7B-cpt & \underline{27.6} & \underline{26.8} & 12.6 & \underline{41.1} & \underline{44.5} & \underline{36.3} & 46.4 & \underline{33.6} \\
13B-ins & 22.1 & 22.7 & \textbf{16.4} &  25.7 & 19.0 &  22.7 &  34.0 & 23.2 \\
\midrule
Proxy Tuning &  25.7 & \underline{26.8} & 13.3 & 33.2 & 39.6 & 33.3 & \underline{47.3} & 31.3 \tiny{(-6.8\%)}\\
\textsc{TriMix} (PPL) &  \textbf{30.4} & \textbf{33.2} & \underline{13.8} & \textbf{46.1} & \textbf{50.8} & \textbf{43.2} & \textbf{53.6} & \textbf{38.7} \tiny{(+15.2\%)}\\
\bottomrule
\end{tabular}
\end{small}
\caption{Evaluation results on Llama2 models. }
\label{tab:llama2}
\end{table*}

\begin{table}[t]
\setlength\tabcolsep{5pt}
\centering
\begin{small}
\begin{tabular}{l|cccc|l}
\toprule
\textbf{Model} & \textbf{bod} & \textbf{uig} & \textbf{kaz} & \textbf{mvf} & \textbf{Average} \\
\midrule
4B-base & 24.2 & 32.0 & 24.7 & 17.2 & 24.5 \\
4B-cpt & 35.7 & 36.0 & 33.2 & 19.1 & 31.0 \\
12B-ins & \underline{49.7} & \textbf{57.6} & \underline{50.8} & 24.1 &  \underline{45.6} \\
\midrule
Proxy Tuning & 49.6 & 54.4 & 48.5 & \underline{24.6} & 44.3 \tiny{(-2.9\%)}\\
\textsc{TriMix} (PPL) & \textbf{54.8} & \underline{55.9} & \textbf{51.8} & \textbf{29.4} & \textbf{48.0} \tiny{(+5.3\%)}\\
\bottomrule
\end{tabular}
\end{small}
\caption{Evaluation results on Gemma3 models.}
\label{tab:gemma3}
\end{table}

\subsection{Main Results}

In Table~\ref{tab:main_experiment}, we report the average performance of different methods on the Qwen2.5 family. See the detailed results on individual tasks in Appendix~\ref{app:full_results}.

\paragraph{Superiority of Tri-Source Fusion}
In the setting that combines 1.5B models with a 3B model, \textsc{TriMix} (PPL) outperforms single-model baselines. It also surpasses other multi-model collaboration methods, including model merging and contrastive decoding, when using the same CPT model size. These results highlight the effectiveness of our approach in leveraging the scaling benefits of larger models.

\begin{figure}[t]
\centering
\includegraphics[width=\linewidth]{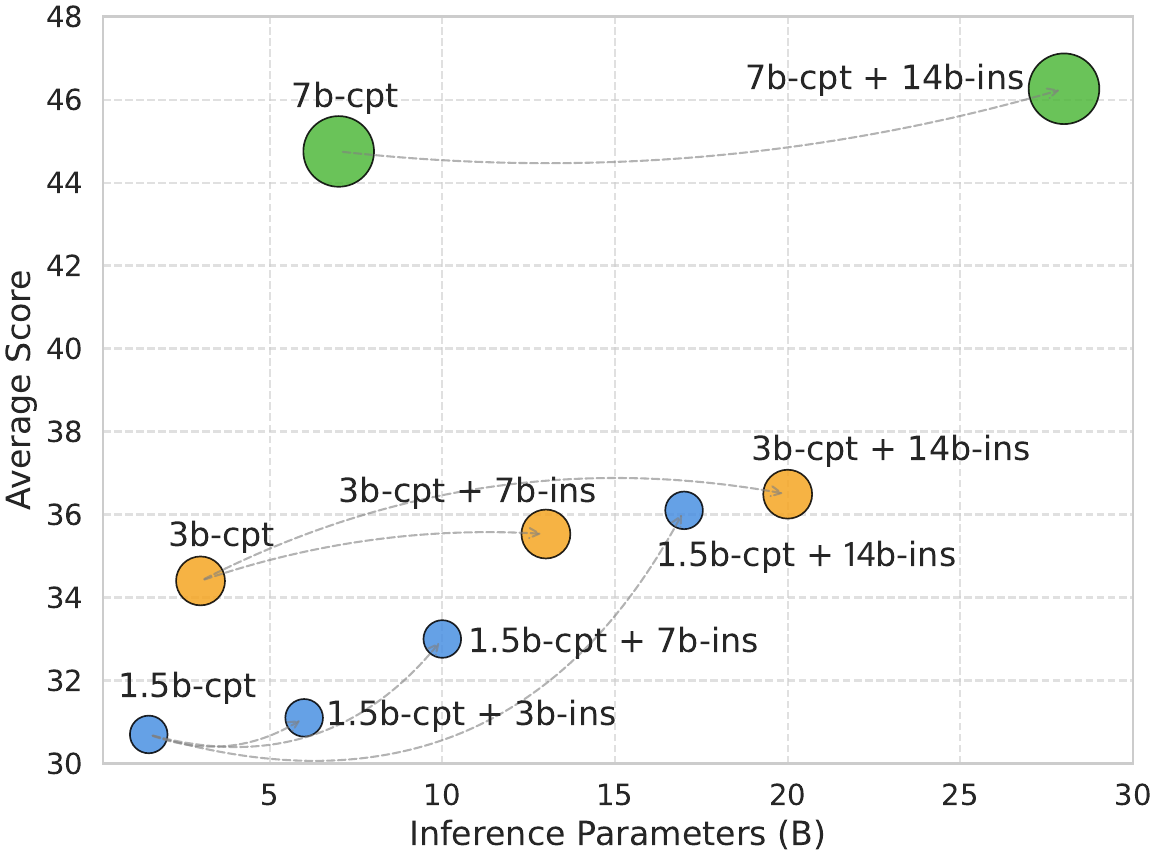}
\caption{Performance of \textsc{TriMix} (PPL) across different combinations of model sizes. Scores are reported on MiLiC-Eval and averaged over four languages. Circle sizes represent the scale of training parameters.}
\label{fig:scale}
\end{figure}

\paragraph{Scalability} 
When scaling the instruction-following model from 3B to 14B (with the CPT model fixed at 1.5B), \textsc{TriMix} (PPL) yields consistent performance gains, achieving a +1.7\% absolute improvement (+4.9\% relative) over the strong 14B-ins baseline. On the other hand, scaling the CPT model, as shown in Figure~\ref{fig:scale}, leads to steady improvements when combined with larger models of various sizes.

In practice, LRL researchers can start with the largest CPT model they can feasibly train and apply \textsc{TriMix} in combination with a larger instruction-tuned model to achieve stable performance gains.
In Appendix~\ref{app:incomplete-cpt}, we additionally analyze the effectiveness of \textsc{TriMix} when the CPT model undergoes only partial training due to computational constraints.

\paragraph{Weighting Strategies}
Although Proxy Tuning is designed to exploit the capacity of larger models, it consistently underperforms in the LRL settings, often yielding worse results than even the standalone CPT model. We will further analyze the underlying reason in Section~\ref{sec:discussion}.

Among our two unsupervised selection strategies, Perplexity-Guided (PPL) selection generally outperforms Entropy-Guided (ENT) selection. We attribute this advantage to the fact that perplexity is computed over the entire input prompt, providing a global contextual signal, whereas entropy depends only on the next-token distribution. Still, a gap remains between our best heuristic and the Upper Bound, indicating that more advanced selection mechanisms could further improve performance.

\subsection{Generalizability}
To validate the generalizability of \textsc{TriMix}, we extend our evaluation to three additional LLM families (Gemma3, Llama2 and Llama3.2) and on four additional low-resource Indian languages.

Table~\ref{tab:gemma3} reports the results on the Gemma3 family. Gemma3-12B-ins is a particularly strong baseline, outperforming all Qwen2.5 configurations of logit fusion. Despite this high starting point, \textsc{TriMix} (combining 4B-cpt and 12B-ins) yields an average relative improvement of +5.3\% over the best single model. This indicates that our method is still effective even when the backbone model is already capable in the target LRL.

We further test on Llama2 (7B-cpt + 13B-ins) covering four additional Indian languages, as shown in Table~\ref{tab:llama2}. \textsc{TriMix} demonstrates strong gains on Indian languages, improving over the 7B-cpt baseline by large margins (e.g., +5.1 on Tamil, +7.2 on Bengali). This further confirms that our framework generalizes effectively across language families. 
We report the results on Llama3.2 in Appendix~\ref{app:full_results}, where \textsc{TriMix} outperforms the single-model baseline by 8.7\% relatively.

We observe that \textsc{TriMix} gains are occasionally marginal or negative for Tibetan (\texttt{bod}) and Mongolian (\texttt{mvf}) in certain configurations. We hypothesize this stems from the extremely low tokenizer fertility for these scripts in English-centric models. As noted by \citet{zhang-etal-2025-milic}, Qwen2.5's tokenization efficiency for Tibetan is approximately $10\times$ lower than for English. 
Future work will investigate vocabulary expansion techniques to mitigate this bottleneck.

\begin{figure*}[t]
\centering
\includegraphics[width=\linewidth]{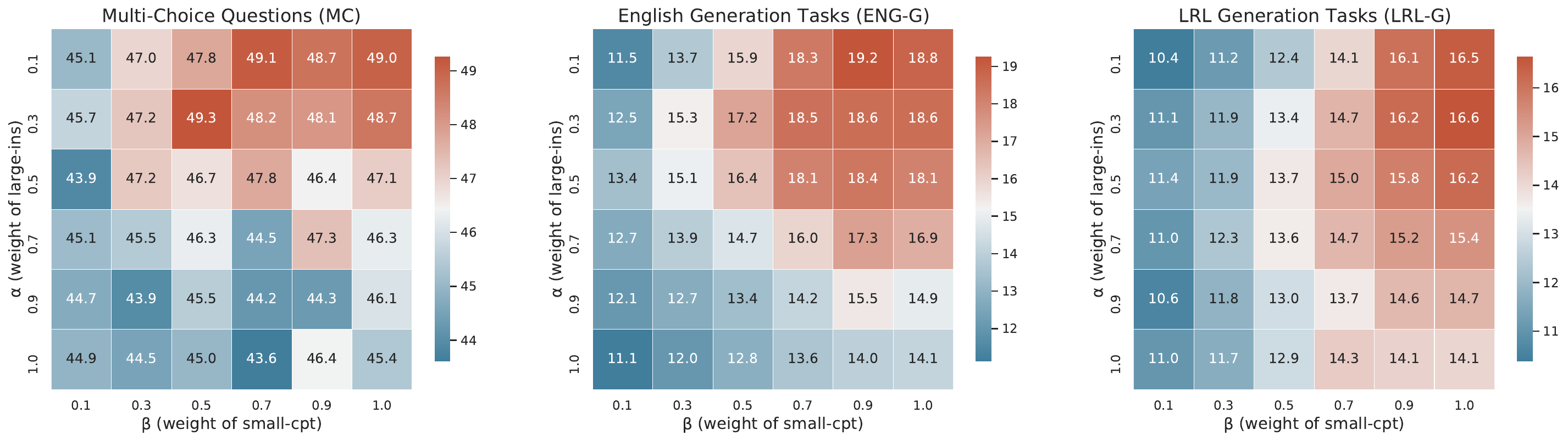}
\caption{The average score across all the tasks on MiLiC-Eval for each pair of hyperparameters in \textsc{TriMix}.}
\label{fig:heatmap}
\end{figure*}

\section{Discussion}
\label{sec:discussion}

To better understand the reason behind the effectiveness of \textsc{TriMix}, we analyze the weights assigned to each model under different hyperparameter selection strategies, and interpret the weighting behaviors through the lens of model divergence from the base model.

\subsection{Analysis of Hyperparameter Selection}

To understand the mechanics of \textsc{TriMix}, we conduct a grid search of the fusion weights $(\alpha, \beta)$ for the Qwen2.5 1.5B-cpt + 3B-ins setting. Figure~\ref{fig:heatmap} visualizes the average performance across all four LRLs on MiLiC-Eval for different hyperparameter combinations.
We observe a consistent pattern across tasks: the optimal region is situated in the high $\beta$, low $\alpha$ quadrant, generally satisfying the condition $\beta > \alpha$. This indicates that significantly greater weight must be assigned to the smaller LRL-adapted CPT model ($L_{\text{small-cpt}}$) than to the large instruction-tuned model ($L_{\text{large-ins}}$). This empirical finding directly contradicts the core assumption of Proxy Tuning (where $\alpha=1$ is by default), providing the potential explanation for its sub-optimal performance in LRL scenarios.

Table~\ref{tab:hyperparameter} reports the mean selected hyperparameters of different strategies, under the Qwen2.5 1.5B-cpt + 3B-ins setting. The perplexity-guided strategy yields a mean configuration ($\alpha$ = 0.11, $\beta$ = 0.91) that closely mirrors the distribution of the Upper Bound ($\alpha$ = 0.33, $\beta$ = 0.74). In contrast, the entropy-guided strategy ($\alpha$ = 0.95, $\beta$ = 0.61) assigns excessively high weight to the large instructed model, similar to the ineffective Proxy Tuning approach. To summarize, the alignment between the PPL heuristic and the upper bound configuration explains its consistent superiority in the main results (Table~\ref{tab:main_experiment}).

\begin{table}[t]
\centering
\begin{small}
\begin{tabular}{l|cc}
\toprule
\textbf{Model} & $\bm{\alpha}$ & $\bm{\beta}$ \\
\midrule
Upper Bound & 0.33$\pm$0.26 & 0.74$\pm$0.29\\
Proxy Tuning & 1.00$\pm$0.00 & 1.00$\pm$0.00 \\
\textsc{TriMix} (ENT) & 0.95$\pm$0.14 & 0.61$\pm$0.40 \\
\textsc{TriMix} (PPL) & 0.11$\pm$0.05 & 0.91$\pm$0.06 \\
\bottomrule
\end{tabular}
\end{small}
\caption{Mean and standard deviation of hyperparameters ($\alpha$, $\beta$) selected by different fusion strategies. }
\label{tab:hyperparameter}
\end{table}

\subsection{Explaining the Balance of Competence}
Previous studies~\cite{jacot2018neural,korbak2022reinforcement} suggest that changes in a model’s output distribution are closely correlated with changes in its parameters. Accordingly, a large divergence between the logits before and after specialized learning indicates that the model has undergone sufficient specialization.
Building on this presumption, we hypothesize that a specialized model should receive a larger fusion weight when it remains closer to the common base model. 
When the component models of logit fusion exhibit asymmetric divergence, their fusion weights should be recalibrated accordingly to balance their specialized contributions during fusion.

\paragraph{Setup}
To validate our hypothesis, we conduct an analysis in two domains: (i) the code domain studied in the original Proxy Tuning work, and (ii) LRLs. We quantify how specialized models differ from the base model by computing the KL divergence between the predicted token distributions of the first generated token under identical inputs.

For the code domain, we consider three models: Qwen2.5-1.5B-base, Qwen2.5-1.5B-coder~\cite{hui2024qwen2coder}, and Qwen2.5-3B-ins. We compute the KL divergence between 1.5B-coder and 1.5B-base to capture the effect of code-domain adaptation, and between 3B-ins and 1.5B-base to measure the effect of instruction tuning combined with  scaling. The measurements are conducted on the HumanEval code completion benchmark~\cite{chen2021codex}, following the original work. 
For LRLs, we use models from the same Qwen2.5 family: Qwen2.5-1.5B-base, Qwen2.5-1.5B-cpt, and Qwen2.5-3B-ins. We compute the KL divergence between 1.5B-cpt and 1.5B-base to measure the effect of continual pretraining on LRL data. 
Evaluation is conducted on MiLiC-Eval, averaging the results over 7 tasks.

\begin{table}[t]
\centering
\begin{small}
\begin{tabular}{l|cc}
\toprule
\textbf{Task} & \textbf{3B-ins} & \textbf{1.5B-coder} \\
\midrule
HumanEval & 0.140 & 0.142 \\
\midrule
\midrule
\textbf{Task} & \textbf{3B-ins} & \textbf{1.5B-cpt} \\
\midrule
MiLiC-Eval (bod) & 1.387 & 0.972 \\
MiLiC-Eval (uig) &  2.124 & 1.080 \\
MiLiC-Eval (kaz) & 2.233 & 1.077 \\
MiLiC-Eval (mvf) & 1.859 & 1.823 \\
\bottomrule
\end{tabular}
\end{small}
\caption{KL divergence of various model variants relative to Qwen2.5-1.5B-base. Scores are calculated using the predicted distributions of the first generated token.}
\label{tab:kl}
\end{table}

\paragraph{Results}
Table~\ref{tab:kl} reports the KL divergence between different specialized models and a shared base model. In the code domain, the KL divergence between 1.5B-coder and 1.5B-base is comparable to that between 3B-ins and 1.5B-base. This suggests that the code CPT trained on over 5T tokens of code induces parameter and distributional changes of a similar magnitude to those introduced by instruction tuning combined with scaling. Under this condition, assigning equal fusion weights (i.e., $\alpha = \beta = 1$) is reasonable to achieve strong performance on code-related tasks, which explains the empirical success of Proxy Tuning in this domain.

In contrast, a markedly different pattern emerges in LRL settings. The KL divergence between 3B-ins and 1.5B-base is generally larger than that between the 1.5B-cpt and 1.5B-base. This indicates that instruction tuning and scaling introduce substantially stronger distributional shifts than LRL CPT. As a result, Proxy Tuning tends to overemphasize the large instruction-tuned model, thereby suppressing the LRL competence learned during CPT. Notably, LRL CPT typically involves fewer than 1B tokens, and the resulting language competence is far from saturated, suggesting that it should receive a larger fusion weight. These observations motivate rebalanced fusion in LRL scenarios, where the LRL component must be up-weighted ($\beta > \alpha$) to compensate for its lower intrinsic divergence from the base model.

Overall, these findings provide empirical support for our hypothesis that optimal fusion weights are closely tied to each component’s divergence from the base model, and that different domains exhibit fundamentally different divergence structures, motivating the adaptive weighting strategy of \textsc{TriMix}.

\subsection{Ablation of Benefit Sources}
To understand how each source of benefit in \textsc{TriMix} affects the overall performance, we conduct an ablation study of the three benefit vectors in Eq.~\ref{eq:decomposition}.
In Table~\ref{tab:ablation}, we report the performance of Qwen2.5 1.5B-cpt + 7B-ins combination on the Kazakh tasks of MiLiC-Eval, using perplexity-based hyperparameter selection.
The largest degradation occurs when removing the language modeling vector ($\delta_{L}$), indicating that language modeling ability is the dominant contributor. The scaling effect ($\delta_{S}$) is the second most important, while the task-solving vector ($\delta_{T}$) provides smaller but consistent gains.
These findings align with our observation that upweighting language ability (high $\beta$) is critical in LRL settings.

\begin{table}[t]
\centering
\begin{small}
\begin{tabular}{ll}
\toprule
\textbf{Setting} & \textbf{Score} \\
\midrule
\textsc{TriMix} & 38.1 \\
\midrule
w/o Task Solving Vector $\delta_{T}$ & 37.2  \tiny(-0.9)\\
w/o Language Modeling Vector $\delta_{L}$ & 30.5 \tiny(-7.6)\\
w/o Scaling Effect Vector $\delta_{S}$ & 33.9 \tiny(-4.2) \\
\bottomrule
\end{tabular}
\end{small}
\caption{Ablation study of the three benefit vectors in \textsc{TriMix}. }
\label{tab:ablation}
\end{table}

\section{Related Works}
\paragraph{Language Adaptation}

Existing approaches for adapting LLMs to LRLs can be broadly categorized into two paradigms:
Parameter-level approaches rely on continual pretraining~\cite{yong-etal-2023-bloom,fujii2024continual,aggarwal-etal-2025-improving,li-etal-2025-group} or fine-tuning~\cite{su-etal-2024-unlocking,singh-etal-2024-aya,shaham-etal-2024-multilingual} on LRL data to improve language competence;
prompt-level approaches teach LLMs new languages by injecting external linguistic resources, such as dictionaries~\cite{zhang-etal-2024-teaching,li-etal-2025-context-learning} and grammar books~\cite{tanzer2024a,hus-anastasopoulos-2024-back,zhang-etal-2025-read,pei-etal-2025-understanding} into the input context.

Beyond adapting a single model, recent work has explored multi-model collaboration to combine capabilities for LRLs. Parameter-level techniques include model merging~\cite{tao-etal-2024-unlocking,huang-etal-2024-chat,cao2025paramdelta,yamaguchi2025adapting}, layer swapping~\cite{bandarkar2025layer}, and model stacking~\cite{huang2024mindmerger,schmidt-etal-2024-self,su-etal-2025-multilingual}. At the prompt level, post-editing methods~\cite{cheng-etal-2024-scale,li-etal-2024-improving-cross,deoghare-etal-2024-together} are widely used. 

In contrast to these approaches, we explore an orthogonal yet underexplored direction for LRL adaptation: logit-level fusion, which is  computationally efficient than parameter updating and provides finer-grained control over model behavior than prompt-level methods.

\paragraph{Logit Fusion}
Logit fusion aims to combine token-level output distributions from multiple models during inference, addressing the challenges in hallucinations~\cite{shi-etal-2024-trusting,kim-etal-2024-adaptive},  reasoning~\cite{o2023contrastive,tao-etal-2025-epicode,zhang2025logit}, and safety~\cite{zhong-etal-2024-rose,fu-etal-2025-unlocking}. One line of works amplifies desirable behaviors of stronger models by contrasting them with weaker ones, typically through contrastive decoding~\cite{li-etal-2023-contrastive,chuang2024dola,yu-etal-2025-prunecd}. The other line combines complementary capabilities across models. This includes transferring specialized abilities learned by smaller models to larger ones~\cite{liu2024tuning,mitchell2024an}, as well as directly ensembling multiple models~\cite{mavromatis2024pack}. To guide fusion, prior work has explored confidence- or uncertainty-based signals~\cite{garces-arias-etal-2024-adaptive,jin-etal-2024-dvd,lee-etal-2025-uncertainty}, divergence-based criteria~\cite{fan2024on}, or learned routers~\cite{shen-etal-2024-learning}.

In multilingual contexts, preliminary efforts have applied logit fusion to mitigate hallucinations in machine translation~\cite{sennrich-etal-2024-mitigating,waldendorf-etal-2024-contrastive} or to improve multilingual mathematical reasoning~\cite{zhu-etal-2024-multilingual-contrastive}.
However, they are primarily tailored for high-resource languages and typically do not account for scaling effects.
In contrast, we systematically analyze the limitations of existing logit fusion strategies in LRL settings and propose a refined framework that balances various sources of benefits.

\section{Conclusion}
We propose \textsc{TriMix}, a logit fusion framework for LRLs that balances language competence, task competence, and scaling benefits while requiring continual pretraining only on a small model. Experiments across four LLM families and eight languages demonstrate its effectiveness and generalizability. Our analysis shows that prioritizing language-specific competence is crucial for LRLs, challenging the large-model-dominant assumption in prior work. We hope that \textsc{TriMix} offers a practical path for LRL adaptation under limited data and computational budgets.

\section*{Limitations}

\paragraph{Effect of Vocabulary Expansion} \textsc{TriMix} does not explicitly examine the effect of vocabulary expansion, nor does it address potential vocabulary or tokenization misalignment across component models. In our setting, all models share a largely compatible tokenizer, which allows for direct logit-level fusion without additional alignment mechanisms. However, for languages with underrepresented scripts, it is often suggested to expand the vocabulary before CPT to improve encoding efficiency. This might result in mismatched vocabularies between component models during logit fusion, requiring additional preprocessing steps such as vocabulary mapping or re-tokenization, which we leave for future work.

\paragraph{Applicability to Closed-Source LLMs} \textsc{TriMix} requires direct access to the output logits of all component models in order to perform logit-level fusion. This assumption restricts its applicability to open-source models or locally deployed systems. For closed-source or API-based models, where only final decoded outputs are available, \textsc{TriMix} cannot be directly applied. Exploring proxy signals or alternative fusion strategies that do not rely on raw logits remains an open challenge.

\paragraph{Inference Overhead} 
While \textsc{TriMix} significantly reduces the computational cost associated with continual pretraining over large models, it introduces additional overhead at inference time. Specifically, inference requires running multiple component models in parallel and performing logit fusion at each generation step. This increases latency and memory consumption compared to single-model inference. 
This overhead can be partially mitigated through techniques such as model quantization. For instance, with INT8 quantization, the quantized \textsc{TriMix} configuration still significantly outperforms the strongest single-model baseline, with less than a 0.5\% performance drop compared to the non-quantized configuration. 
Future work could investigate more sophisticated techniques like model compression, selective activation, or distillation-based variants of \textsc{TriMix} to mitigate this overhead.

\paragraph{Potential Risks} 
Although \textsc{TriMix} improves performance on many LRL benchmarks, it may still produce incorrect or nonsensical outputs in these languages and should therefore be used with caution.

\section*{Acknowledgements}
This work is supported in part by Beijing Natural Science Foundation (L253001) and Natural Science Foundation of China (92570207). We thank the anonymous reviewers for their valuable feedback. 
For any correspondence, please contact Yansong Feng.

\bibliography{anthology,custom}

\appendix

\section{Implementation Details}
\label{app:implementation}

\subsection{Selected Languages}
\label{app:languages}
Table~\ref{tab:language_info} summarizes the linguistic details and speaker populations of the target languages, which collectively have over 475 million speakers but receive limited support from existing LLMs.

\subsection{Continual Pretraining}
\label{app:cpt}
\paragraph{Data}
For continual pretraining of Qwen2.5-1.5B, Gemma3-4B, Llama3.2-1B on the minority languages in China, we use the MC$^2$ corpus~\cite{zhang-etal-2024-mc2}. To mitigate catastrophic forgetting of English capabilities, we additionally incorporate English data from C4~\cite{dodge-etal-2021-documenting}, amounting to 20\% of the size of the target-language corpus.

\paragraph{Hyperparameters}
We perform continual pretraining using the Megatron framework~\cite{shoeybi2019megatron}. The models are trained for one epoch with a batch size of 1M tokens, a learning rate of $2\times10^{-5}$, and a warmup ratio of 0.01 on eight L40S GPUs.

\subsection{Baselines}

\paragraph{Model Merging}
For model merging, we use Arcee’s MergeKit~\cite{goddard-etal-2024-arcees}. For the CPT model, the density is set to 1.0 and the weight to 0.2. For the instruction-tuned model, the density is set to 0.2 and the weight to 0.8.

\paragraph{Logit Arithmetic}
For Proxy Tuning, we use the official implementation provided by \citet{liu2024tuning}\footnote{\url{https://github.com/alisawuffles/proxy-tuning}}. We adapt the code to support contrastive decoding in our experimental setup. For the contrsative decoding, we set $\beta$ to 0.5 and set the plausibility threshold to 0.1, following the original paper.

\subsection{Evaluation}
\paragraph{Data}
In MiLiC-Eval, both topic classification and machine translation consist of two subsets. We use the \textit{Passage} subset for the topic classification task and the \textit{Dialogue} subset for the machine translation task.
For title generation, since the inputs are long and evaluation is computationally expensive, we randomly sample 200 instances for evaluation. We report the test instance counts per language and corresponding metrics in Table~\ref{tab:metric}.
We use INT8 quantization to reduce GPU memory consumption during inference.

Our use of these existing artifacts is solely for evaluating multilingual models and is consistent with their intended purposes.

\begin{table}[t]
\small
\centering
\setlength\tabcolsep{3.5pt}
\begin{tabular}{l|ccc}
\toprule
\textbf{Name} & \textbf{Family}  & \textbf{Script} & \textbf{Population} \\
\midrule
Tibetan (bod) &  Sino-Tibetan & Tibetan & 7M \\
Uyghur (uig) &  Turkic & Arabic & 12M \\
Kazakh (kaz) & Turkic & Arabic & 1.6M \\
Mongolian (mvf) &  Mongolic &  Mongolian & 6M \\
Tamil (tam) & Dravidian & Tamil & 79M \\
Telugu (tel) & Dravidian & Telugu & 96M \\
Odia (ory) & Indo-Euro. & Odia & 35M \\
Bengali (ben) &  Indo-Euro. & Bengali & 240M \\
\bottomrule
\end{tabular}
\caption{Language families, writing systems, and populations of the LRLs in our study. }
\label{tab:language_info}
\end{table}

\begin{table}[t]
\centering
\begin{small}
\begin{tabular}{lcc}
\toprule
\textbf{Dataset} & \textbf{Size} & \textbf{Metric} \\
\midrule
\textit{MiLiC-Eval} \\
Topic Classification & 600 & Accuracy \\
Response Selection & 507 & Accuracy \\
Reading Comprehension & 250 & Accuracy \\
Title Generation & 200 & ROUGE-L \\
English-to-LRL Translation & 773 & chrF++ \\
LRL-to-English Translation & 773 & chrF++ \\
Math Reasoning & 250 & Accuracy \\
\midrule
Belebele & 900 & Accuracy \\
SIB-200 & 204 & Accuracy \\
\bottomrule
\end{tabular}
\end{small}
\caption{Summary of evaluation datasets with test instance counts per language and corresponding metrics.}
\label{tab:metric}
\end{table}

\paragraph{Hyperparameters}
We adopt a 5-shot in-context learning setting during evaluation. When the input exceeds the maximum context length of a model, we reduce the number of in-context examples accordingly.
For each experimental setting, we evaluate up to 28 datasets (4 languages $\times$ 7 tasks). To improve efficiency, each setting is evaluated only once. To ensure fairness and comparability across settings, we use the same random seed throughout all experiments.

\section{Additional Results}
\subsection{Larger Hyperparameter Search Range}
\label{app:hyperparam_search}
In our experiments, we restrict the hyperparameter search range to $[0,1]$. When extending this range, we observe that model performance degrades substantially as the logit scales of different component models become increasingly mismatched.
For example, in the setting of Qwen2.5 1.5B-cpt + 3B-ins, Table~\ref{tab:larger_hyperparameter} reports the performance on the Kazakh reading comprehension task from MiLiC-Eval under different hyperparameter values $\{0.1, 1, 5, 10\}$. When either fusion weight becomes large, performance drops sharply, falling below 5\% in extreme cases.

\begin{table}[t]
\centering
\begin{small}
\begin{tabular}{l|rrrr}
\toprule
\diagbox{$\bm{\alpha}$}{$\bm{\beta}$} & \textbf{0.1} & \textbf{1} & \textbf{5} & \textbf{10} \\
\midrule
\textbf{0.1} & 52.5 & 54.5 & 3.0 & 3.5\\
\textbf{1} &  50.5 & 54.0 & 13.0 & 6.0\\
\textbf{5} & 48.0 & 48.5 & 48.0 & 18.5 \\
\textbf{10} & 48.5 & 48.0 & 49.5 & 36.0 \\
\bottomrule
\end{tabular}
\end{small}
\caption{Accuracy (\%) on the Kazakh reading comprehension task from MiLiC-Eval, using a larger range of hyperparameters, in the setting of Qwen2.5-1.5B-cpt + Qwen2.5-3B-ins.}
\label{tab:larger_hyperparameter}
\end{table}

\subsection{Full Evaluation Results}
\label{app:full_results}

\paragraph{Llama2}
In Table~\ref{tab:llama-indian-full}, we report the results of Llama2 on SIB-200 and Belebele in the four Indian languages.

\paragraph{Llama3.2}
In Table~\ref{tab:llama3.2}, we report the evaluation results on the Llama3.2 series. \textsc{TriMix} improves over the strongest single model by +2.2 points (8.7\% relative). 

\paragraph{Qwen2.5} 
We report the full results of the Qwen2.5 series on the MiLiC-Eval benchmark: Table~\ref{tab:main_experiment_bod} for Tibetan, Table~\ref{tab:main_experiment_uig} for Uyghur, Table~\ref{tab:main_experiment_kaz} for Kazakh, and Table~\ref{tab:main_experiment_mvf} for Mongolian.

\subsection{Fusion After Inadequate CPT}
\label{app:incomplete-cpt}
In practice, computational constraints often prevent full continual pretraining (CPT). We therefore examine whether \textsc{TriMix} remains effective when applied to partially trained models.

To simulate a limited-budget CPT setting, we conduct experiments with Qwen2.5-7B and evaluate performance on seven Kazakh tasks from MiLiC-Eval. As shown in Table~\ref{tab:incomplete-cpt}, even after only one quarter of the full CPT steps, \textsc{TriMix} yields a relative improvement of 5.4\%.

Importantly, TriMix acts as a multiplier rather than an alternative for CPT. Even partially trained models benefit when combined with a stronger instruction-tuned model. This demonstrates that practitioners can maximize limited CPT budgets by leveraging inference-time combination with a larger off-the-shelf model.

\subsection{Case Study}
\label{app:case_sudy}
Table~\ref{tab:case_study} presents a qualitative case study comparing different logit fusion methods.
For a prompt that requires generating a story title in Uyghur, Proxy Tuning produces invalid byte sequences that cannot be decoded into meaningful UTF-8 characters. This failure suggests that Proxy Tuning’s large-model-dominant assumption allows weak LRL representations in the large model to overwhelm the contribution of the smaller continual-pretrained model.
In contrast, \textsc{TriMix} well balances the contributions of different models and generates a plausible Uyghur title, demonstrating its effectiveness in LRL settings.

\begin{table}[t]
\centering
\begin{small}
\begin{tabular}{l|ccc|l}
\toprule
\textbf{Model} & \textbf{bod} & \textbf{uig} & \textbf{kaz} & \textbf{Average} \\
\midrule
1B-base & 16.2 & 17.3 & 15.5 & 16.3 \\
1B-cpt & 16.8 & 22.4 & 21.2 & 20.2 \\
3B-ins & 19.6 & 30.8 & 25.1 & 25.2\\
\midrule
\textsc{TriMix} (PPL) & 20.9 & 33.2 & 28.1 & 27.4 \tiny{(+8.7\%)}\\
\bottomrule
\end{tabular}
\end{small}
\caption{Evaluation results on Llama3.2 models. }
\label{tab:llama3.2}
\end{table}

\begin{table}[t]
\centering
\begin{small}
\begin{tabular}{l|cc|c}
\toprule
\textbf{Model} & \textbf{SIB-200} & \textbf{Belebele} & \textbf{Avg.} \\
\midrule
\multicolumn{4}{c}{Tamil} \\
\midrule
7B-base & 27.3 & 25.4 & 26.4\\
7B-cpt & 53.5 & 28.6 & 41.1\\
13B-ins & 26.3 & 25.0 & 25.6\\
Proxy Tuning & 32.3 & 34.1 & 33.2 \\
\textsc{TriMix} (PPL) & 56.6 & 32.6 & 46.1 \\
\midrule
\multicolumn{4}{c}{Telugu} \\
\midrule
7B-base & 18.2 & 25.3 & 21.8 \\
7B-cpt & 64.6 & 24.2 & 44.5 \\
13B-ins & 24.2 & 13.7 & 19.0 \\
Proxy Tuning  & 50.5 & 28.7 & 39.6 \\
\textsc{TriMix} (PPL) & 71.7 & 29.9 & 50.8 \\
\midrule
\multicolumn{4}{c}{Odia} \\
\midrule
7B-base & 18.2 & 20.3 & 19.3 \\
7B-cpt & 52.5 & 20.1 & 36.3 \\
13B-ins & 25.3 & 20.1 & 22.7\\
Proxy Tuning & 41.4 & 25.1 & 33.3 \\
\textsc{TriMix} (PPL) & 60.6 & 25.7 & 43.2 \\
\midrule
\multicolumn{4}{c}{Bengali} \\
\midrule
7B-base & 36.4 & 26.9 & 31.7 \\
7B-cpt & 66.7 & 26.1 & 46.4 \\
13B-ins & 40.4 & 27.6 & 34.0 \\
Proxy Tuning & 61.6 & 33.1 & 47.3 \\
\textsc{TriMix} (PPL) & 74.7 & 32.4 & 53.6 \\

\bottomrule
\end{tabular}
\end{small}
\caption{Accuracy (\%) of different methods on SIB-200 and Belebele, using the Llama2 models. }
\label{tab:llama-indian-full}
\end{table}

\begin{table}[t]
\centering
\begin{small}
\begin{tabular}{ll}
\toprule
\textbf{Model} &  \textbf{Score} \\
\midrule
14B-ins & 34.6 \\
7B-cpt (1/8 steps) & 38.1 \\
7B-cpt (1/4 steps) & 42.1 \\
\midrule
\textsc{TriMix} (1/8 steps) & 41.3 \tiny{(8.4\%)} \\
\textsc{TriMix} (1/4 steps) & 44.2 \tiny{(5.0\%)} \\ 
\bottomrule
\end{tabular}
\end{small}
\caption{Performance of \textsc{TriMix} using checkpoints of incomplete CPT. The numbers in parentheses represent relative improvements over the best single-model baseline.}
\label{tab:incomplete-cpt}
\end{table}

\begin{table}[t]
\small
\centering
\begin{tabular}{p{0.9\columnwidth}}
\toprule

\textbf{Input:} Please write a title for the following article in Uyghur:\\ \textit{\{A news story about Messi in Uyghur.\}}\\

\midrule
\textbf{Gold:} \\
\includegraphics[scale=0.45]{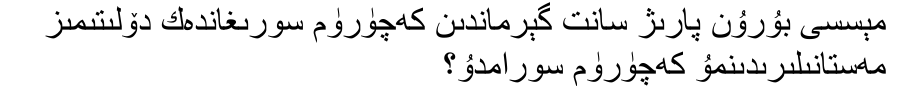} \\
\textit{Will Messi apologize to our fans, just like he apologized to Paris Saint-Germain?} \\
\midrule
\textbf{Proxy Tuning:} \\
\includegraphics[scale=0.45]{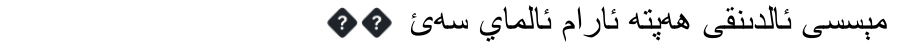} \\
\textit{Messi didn't rest last week [unreadable characters]} \\
\midrule
\textbf{TriMix:} \\
\includegraphics[scale=0.45]{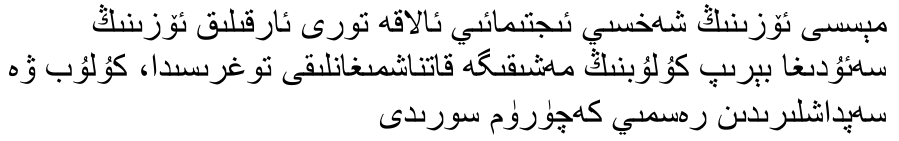} \\
\textit{Messi officially apologized to the club and his teammates via his personal social media regarding his absence from club training due to his trip to Saudi Arabia.} \\
\bottomrule
\end{tabular}
\caption{Case study of different methods on the title generation task in Uyghur, under the setting of Llama2 7B-cpt + 13B-ins.}
\label{tab:case_study}
\end{table}

\begin{table*}[t]
\small
\centering
\begin{tabular}{l|ccccccc|c}
\toprule
\textbf{Model} &  \textbf{{Topic}} & \textbf{{Read.}} & \textbf{Resp.} & \textbf{Title} & \textbf{MT} & \textbf{MT} & \textbf{Math} & \textbf{Avg.}\\
 & \textbf{CLS} & \textbf{Comp.} & \textbf{Sel.} & \textbf{Gen.} & xx2en & en2xx &  & \\
\midrule
Qwen2.5-1.5B-base & 28.9 & 43.8 & 35.1 & 21.6 & 8.4 & 8.8 & 9.3 & 22.3 \\
Qwen2.5-1.5B-cpt & 48.9 & 45.3 & 30.8 & 24.1 & 11.2 & 14.1 & 5.5 & 25.7 \\
Qwen2.5-1.5B-ins & 28.1 & 34.3 & 33.3 & 13.5 & 10.6 & 11.9 & 6.5 & 19.8 \\
Contrastive Decoding & 61.1 & 40.5 & 26.5 & 18.4 & 13.3 & 14.4 & 0.5 & 25.0 \\
Model Merging & 49.8 & 43.0 & 36.3 & 27.6 & 12.1 & 12.5 & 7.8 & 27.0 \\
\midrule
Qwen2.5-3B-ins & 34.1 & 42.5 & 39.0 & 24.1 & 12.9 & 12.5 & 4.5 & 24.2 \\
Proxy Tuning (1.5B-cpt + 3B-ins) & 33.9 & 43.0 & 39.1 & 18.8 & 14.0 & 13.3 & 2.5 & 23.5 \\
\textsc{TriMix} (ENT) (1.5B-cpt + 3B-ins) & 38.9 & 43.0 & 37.8 & 19.9 & 11.9 & 14.4 & 2.5 & 24.1 \\
\textsc{TriMix} (PPL) (1.5B-cpt + 3B-ins) & 43.9 & 43.0 & 36.9 & 23.6 & 12.5 & 14.3 & 5.0 & 25.6 \\
\textsc{TriMix} (Upper Bound) (1.5B-cpt + 3B-ins) & 46.0 & 50.0 & 40.5 & 24.0 & 14.0 & 14.4 & 6.5 & 27.9 \\
\midrule
Qwen2.5-7B-ins & 39.4 & 48.3 & 39.6 & 29.4 & 14.2 & 13.2 & 13.3 & 28.2 \\
Proxy Tuning (1.5B-cpt + 7B-ins) & 46.8 & 45.0 & 35.1 & 22.2 & 15.7 & 13.6 & 5.5 & 26.3 \\
\textsc{TriMix} (ENT) (1.5B-cpt + 7B-ins) & 45.4 & 44.5 & 37.1 & 24.4 & 15.7 & 14.3 & 5.5 & 26.7 \\
\textsc{TriMix} (PPL) (1.5B-cpt + 7B-ins) & 58.7 & 44.5 & 38.1 & 22.3 & 12.8 & 14.1 & 3.5 & 27.7 \\
\midrule
Qwen2.5-14B-ins & 74.8 & 53.0 & 41.0 & 29.9 & 20.9 & 15.0 & 15.0 & 35.7 \\
Proxy Tuning (1.5B-cpt + 14B-ins) & 77.0 & 54.5 & 38.6 & 21.5 & 21.3 & 15.8 & 3.0 & 33.1 \\
\textsc{TriMix} (ENT) (1.5B-cpt + 14B-ins) & 74.0 & 54.5 & 40.1 & 21.5 & 21.3 & 16.4 & 5.0 & 33.3 \\
\textsc{TriMix} (PPL) (1.5B-cpt + 14B-ins) & 84.5 & 50.5 & 38.6 & 21.8 & 18.0 & 15.9 & 5.5 & 33.6 \\
\bottomrule
\end{tabular}
\caption{Scores (\%) of different methods on the \textbf{Tibetan} tasks of MiLiC-Eval, using the Qwen2.5 models. 
\textbf{Topic CLS} refers to topic classification.
\textbf{Read. Comp.} refers to Reading Comprehension.
\textbf{Resp. Sel.} refers to Response Selection.
\textbf{Title Gen.} refers to Title Generation.
\textbf{MT} refers to Machine Translation.
xx2en denotes translation from LRLs to English.
en2xx denotes translation from English to LRLs.
\textbf{Math} refers to Math Reasoning.
}
\label{tab:main_experiment_bod}
\end{table*}

\begin{table*}[t]
\small
\centering
\begin{tabular}{l|ccccccc|c}
\toprule
\textbf{Model} &  \textbf{{Topic}} & \textbf{{Read.}} & \textbf{Resp.} & \textbf{Title} & \textbf{MT} & \textbf{MT} & \textbf{Math} & \textbf{Avg.}\\
 & \textbf{CLS} & \textbf{Comp.} & \textbf{Sel.} & \textbf{Gen.} & xx2en & en2xx &  & \\
\midrule
Qwen2.5-1.5B-base & 47.5 & 48.7 & 38.7 & 15.3 & 13.1 & 6.6 & 16.2 & 26.6 \\
Qwen2.5-1.5B-cpt & 81.3 & 65.3 & 44.8 & 16.8 & 37.2 & 20.5 & 25.7 & 41.7 \\
Qwen2.5-1.5B-ins & 39.0 & 44.2 & 36.4 & 14.4 & 14.1 & 9.4 & 12.3 & 24.3 \\
Contrastive Decoding & 72.2 & 59.0 & 40.8 & 15.6 & 37.2 & 19.9 & 19.5 & 37.7 \\
Model Merging & 83.7 & 61.5 & 46.0 & 18.4 & 38.4 & 20.9 & 24.0 & 41.8 \\
\midrule
Qwen2.5-3B-ins & 71.4 & 38.2 & 39.6 & 17.7 & 19.9 & 10.6 & 15.5 & 30.4 \\
Proxy Tuning (1.5B-cpt + 3B-ins) & 73.6 & 51.0 & 43.2 & 20.4 & 28.5 & 15.0 & 20.0 & 36.0 \\
\textsc{TriMix} (ENT) (1.5B-cpt + 3B-ins) & 65.5 & 48.5 & 44.0 & 20.3 & 28.5 & 16.6 & 13.5 & 33.8 \\
\textsc{TriMix} (PPL) (1.5B-cpt + 3B-ins) & 81.9 & 59.5 & 46.2 & 15.2 & 37.3 & 19.6 & 30.0 & 41.4 \\
\textsc{TriMix} (Upper Bound) (1.5B-cpt + 3B-ins) & 85.3 & 59.5 & 47.4 & 22.1 & 37.3 & 20.9 & 34.0 & 43.8 \\
\midrule
Qwen2.5-7B-ins & 85.1 & 57.0 & 47.8 & 18.2 & 26.4 & 11.7 & 38.0 & 40.6 \\
Proxy Tuning (1.5B-cpt + 7B-ins) & 89.3 & 58.0 & 47.9 & 14.4 & 32.2 & 13.1 & 15.5 & 38.6 \\
\textsc{TriMix} (ENT) (1.5B-cpt + 7B-ins) & 90.7 & 59.0 & 48.4 & 18.9 & 32.2 & 18.9 & 15.5 & 40.5 \\
\textsc{TriMix} (PPL) (1.5B-cpt + 7B-ins) & 83.7 & 62.5 & 50.9 & 15.2 & 38.2 & 17.6 & 30.5 & 42.6 \\
\midrule
Qwen2.5-14B-ins & 89.9 & 61.5 & 57.7 & 22.4 & 30.1 & 13.0 & 25.7 & 42.9 \\
Proxy Tuning (1.5B-cpt + 14B-ins) & 91.5 & 55.5 & 53.1 & 22.1 & 36.3 & 17.2 & 6.5 & 40.3 \\
\textsc{TriMix} (ENT) (1.5B-cpt + 14B-ins) & 95.0 & 55.5 & 53.1 & 22.1 & 36.3 & 19.6 & 7.5 & 41.3 \\
\textsc{TriMix} (PPL) (1.5B-cpt + 14B-ins) & 92.9 & 66.0 & 55.5 & 17.1 & 38.3 & 21.3 & 33.0 & 46.3 \\
\bottomrule
\end{tabular}
\caption{Scores (\%) of different methods on the \textbf{Uyghur} tasks of MiLiC-Eval, using the Qwen2.5 models. 
\textbf{Topic CLS} refers to topic classification.
\textbf{Read. Comp.} refers to Reading Comprehension.
\textbf{Resp. Sel.} refers to Response Selection.
\textbf{Title Gen.} refers to Title Generation.
\textbf{MT} refers to Machine Translation.
xx2en denotes translation from LRLs to English.
en2xx denotes translation from English to LRLs.
\textbf{Math} refers to Math Reasoning.
}
\label{tab:main_experiment_uig}
\end{table*}

\begin{table*}[t]
\small
\centering
\begin{tabular}{l|ccccccc|c}
\toprule
\textbf{Model} &  \textbf{{Topic}} & \textbf{{Read.}} & \textbf{Resp.} & \textbf{Title} & \textbf{MT} & \textbf{MT} & \textbf{Math} & \textbf{Avg.}\\
 & \textbf{CLS} & \textbf{Comp.} & \textbf{Sel.} & \textbf{Gen.} & xx2en & en2xx &  & \\
\midrule
Qwen2.5-1.5B-base & 54.2 & 46.5 & 35.6 & 16.9 & 11.0 & 5.8 & 8.7 & 25.5 \\
Qwen2.5-1.5B-cpt & 57.3 & 52.2 & 42.4 & 23.0 & 27.8 & 14.5 & 19.2 & 33.8 \\
Qwen2.5-1.5B-ins & 47.7 & 38.8 & 34.0 & 14.7 & 13.6 & 8.8 & 6.2 & 23.4 \\
Contrastive Decoding & 55.6 & 51.5 & 40.8 & 16.4 & 29.4 & 16.3 & 14.5 & 32.0 \\
Model Merging & 53.5 & 53.2 & 42.9 & 22.9 & 28.3 & 13.5 & 22.5 & 33.8 \\
\midrule
Qwen2.5-3B-ins & 59.1 & 38.5 & 38.2 & 2.2 & 16.7 & 2.9 & 4.7 & 23.2 \\
Proxy Tuning (1.5B-cpt + 3B-ins) & 54.4 & 54.0 & 41.3 & 15.0 & 22.9 & 12.5 & 6.0 & 29.4 \\
\textsc{TriMix} (ENT) (1.5B-cpt + 3B-ins) & 58.7 & 50.5 & 37.1 & 15.0 & 22.9 & 13.3 & 7.5 & 29.3 \\
\textsc{TriMix} (PPL) (1.5B-cpt + 3B-ins)& 65.3 & 54.5 & 46.4 & 22.6 & 28.6 & 14.0 & 23.0 & 36.4 \\
\textsc{TriMix} (Upper Bound) (1.5B-cpt + 3B-ins) & 68.8 & 55.0 & 47.4 & 23.0 & 28.6 & 14.6 & 24.0 & 37.4 \\
\midrule
Qwen2.5-7B-ins & 64.8 & 51.7 & 40.1 & 13.3 & 21.0 & 1.1 & 21.3 & 30.5 \\
Proxy Tuning (1.5B-cpt + 7B-ins) & 67.1 & 53.5 & 38.6 & 21.3 & 26.0 & 13.6 & 16.0 & 33.7 \\
\textsc{TriMix} (ENT) (1.5B-cpt + 7B-ins) & 74.4 & 53.5 & 38.6 & 21.3 & 26.0 & 13.6 & 16.0 & 34.8 \\
\textsc{TriMix} (PPL) (1.5B-cpt + 7B-ins) & 76.0 & 58.0 & 45.5 & 21.0 & 29.6 & 13.7 & 23.0 & 38.1 \\
\midrule
Qwen2.5-14B-ins & 69.4 & 60.0 & 47.7 & 13.7 & 24.7 & 5.7 & 21.0 & 34.6 \\
Proxy Tuning (1.5B-cpt + 14B-ins) & 79.4 & 56.5 & 46.9 & 24.8 & 29.5 & 14.7 & 9.0 & 37.2 \\
\textsc{TriMix} (ENT) (1.5B-cpt + 14B-ins) & 77.6 & 56.0 & 46.0 & 20.8 & 29.5 & 15.3 & 5.5 & 35.8 \\
\textsc{TriMix} (PPL) (1.5B-cpt + 14B-ins) & 76.8 & 57.5 & 49.1 & 20.7 & 30.3 & 14.7 & 21.0 & 38.6 \\
\bottomrule
\end{tabular}
\caption{Scores (\%) of different methods on the \textbf{Kazakh} tasks of MiLiC-Eval, using the Qwen2.5 models. 
\textbf{Topic CLS} refers to topic classification.
\textbf{Read. Comp.} refers to Reading Comprehension.
\textbf{Resp. Sel.} refers to Response Selection.
\textbf{Title Gen.} refers to Title Generation.
\textbf{MT} refers to Machine Translation.
xx2en denotes translation from LRLs to English.
en2xx denotes translation from English to LRLs.
\textbf{Math} refers to Math Reasoning.
}
\label{tab:main_experiment_kaz}
\end{table*}

\begin{table*}[t]
\small
\centering
\begin{tabular}{l|ccccccc|c}
\toprule
\textbf{Model} &  \textbf{{Topic}} & \textbf{{Read.}} & \textbf{Resp.} & \textbf{Title} & \textbf{MT} & \textbf{MT} & \textbf{Math} & \textbf{Avg.}\\
 & \textbf{CLS} & \textbf{Comp.} & \textbf{Sel.} & \textbf{Gen.} & xx2en & en2xx &  & \\
\midrule
Qwen2.5-1.5B-base & 32.6 & 38.5 & 32.7 & 10.1 & 10.1 & 7.0 & 9.3 & 20.0 \\
Qwen2.5-1.5B-cpt & 39.8 & 42.7 & 28.7 & 15.1 & 10.3 & 9.8 & 4.2 & 21.5 \\
Qwen2.5-1.5B-ins & 36.6 & 32.7 & 31.5 & 3.9 & 12.6 & 7.3 & 7.3 & 18.9 \\
Contrastive Decoding & 33.1 & 37.0 & 24.8 & 10.8 & 10.1 & 9.6 & 1.0 & 18.1 \\
Model Merging & 29.4 & 46.5 & 30.5 & 14.7 & 9.9 & 9.8 & 5.8 & 20.9 \\
\midrule
Qwen2.5-3B-ins & 37.6 & 35.3 & 35.5 & 10.2 & 13.4 & 6.3 & 10.0 & 21.2 \\
Proxy Tuning (1.5B-cpt + 3B-ins) & 33.5 & 42.0 & 35.9 & 8.3 & 14.2 & 9.3 & 5.0 & 21.2 \\
\textsc{TriMix} (ENT) (1.5B-cpt + 3B-ins) & 35.9 & 42.0 & 38.3 & 8.3 & 14.2 & 10.5 & 8.5 & 22.5 \\
\textsc{TriMix} (PPL) (1.5B-cpt + 3B-ins)& 31.9 & 42.0 & 33.4 & 11.2 & 11.5 & 9.9 & 8.0 & 21.1 \\
\textsc{TriMix} (Upper Bound) (1.5B-cpt + 3B-ins) & 43.5 & 45.5 & 40.1 & 11.3 & 14.2 & 10.7 & 12.0 & 25.3 \\
\midrule
Qwen2.5-7B-ins & 37.8 & 44.0 & 40.4 & 9.2 & 14.1 & 4.0 & 11.2 & 22.9 \\
Proxy Tuning (1.5B-cpt + 7B-ins) & 41.9 & 42.5 & 39.8 & 6.7 & 15.2 & 1.2 & 4.0 & 21.6 \\
\textsc{TriMix} (ENT) (1.5B-cpt + 7B-ins) & 43.1 & 41.5 & 38.3 & 9.9 & 15.2 & 9.5 & 3.0 & 22.9 \\
\textsc{TriMix} (PPL) (1.5B-cpt + 7B-ins) & 38.3 & 46.0 & 38.8 & 12.0 & 12.2 & 9.8 & 8.0 & 23.6 \\
\midrule
Qwen2.5-14B-ins & 41.7 & 45.0 & 43.7 & 10.4 & 15.6 & 0.7 & 15.3 & 24.6 \\
Proxy Tuning (1.5B-cpt + 14B-ins) & 51.4 & 42.5 & 45.0 & 10.6 & 16.2 & 7.6 & 1.5 & 25.0 \\
\textsc{TriMix} (ENT) (1.5B-cpt + 14B-ins) & 52.6 & 44.0 & 45.5 & 10.4 & 16.2 & 9.2 & 3.0 & 25.8 \\
\textsc{TriMix} (PPL) (1.5B-cpt + 14B-ins) & 55.8 & 47.5 & 38.8 & 12.7 & 11.6 & 9.9 & 5.5 & 25.9 \\
\bottomrule
\end{tabular}
\caption{Scores (\%) of different methods on the \textbf{Mongolian} tasks of MiLiC-Eval, using the Qwen2.5 models. 
\textbf{Topic CLS} refers to topic classification.
\textbf{Read. Comp.} refers to Reading Comprehension.
\textbf{Resp. Sel.} refers to Response Selection.
\textbf{Title Gen.} refers to Title Generation.
\textbf{MT} refers to Machine Translation.
xx2en denotes translation from LRLs to English.
en2xx denotes translation from English to LRLs.
\textbf{Math} refers to Math Reasoning.
}
\label{tab:main_experiment_mvf}
\end{table*}

\end{document}